\documentclass[11pt]{article}
\usepackage[utf8]{inputenc}
\usepackage[T1]{fontenc}
\usepackage{lmodern}
\usepackage[margin=1in]{geometry}
\usepackage{amsmath,amssymb}
\usepackage{graphicx}
\usepackage{booktabs}
\usepackage{hyperref}
\usepackage{xcolor}
\usepackage{natbib}
\usepackage{enumitem}
\usepackage{placeins}
\usepackage{url}
\usepackage{microtype}
\usepackage{tikz}
\usetikzlibrary{positioning,arrows.meta}

\hypersetup{colorlinks=true, linkcolor=blue, citecolor=blue, urlcolor=blue}

\title{Knowledge-Graph Grounding Helps LLMs Only for Out-of-Training Knowledge:\\
A Controlled Study on Clinical Question Answering}
\author{Madhulatha Mandarapu\thanks{madhulatha@samyama.ai} \and Sandeep Kunkunuru\thanks{sandeep@samyama.ai}}
\date{}

\begin{document}
\maketitle

\begin{abstract}
A recent \emph{Nature Medicine} study reports that general-purpose frontier LLMs outperform specialized
retrieval-augmented clinical tools on medical benchmarks, and that retrieval can \emph{hurt} strong models.
We ask the natural follow-up: does \emph{structured} knowledge-graph (KG) grounding change this, and
\emph{when} does grounding help at all? We contribute two results. First, a reproduction: the study's
headline HealthBench score ($\sim$88) is the \emph{Consensus} variant, not full HealthBench, where
frontier models and ideal completions both score $\sim$46--47 under a physician-calibrated grader
(agreement $82.5\%$); we reproduce GPT-5.2 Consensus $=90.9$ and flag a score-deflating grader bug.
Second, a knowledge-boundary result. Using a graph+vector engine (samyama-graph) over the public
biomedical KG PrimeKG, neither naive triple retrieval nor an agentic natural-language-to-Cypher loop
(82\% successful queries) improves MedQA across a weak$\to$strong model ladder (all $|\Delta| \le 3.4$).
On a \emph{synthetic counterfactual} KG, and on a \emph{hybrid} benchmark mixing known and novel facts,
the identical pipeline lifts out-of-training accuracy from chance to $\sim$100\% ($+68$ to $+79$) while
adding nothing on known facts (a no-LLM arm answers both). Across three regimes (no-knowledge,
graph-aided, hybrid), grounding helps only insofar as the decisive fact lies outside the model's
training---public-KG facts are redundant, private and novel data are where it pays---matching the study's
institutional-data caveat.
\end{abstract}

\section{Introduction}
\citet{vishwanath2026general} report that three frontier LLMs outperform two deployed clinical
retrieval-augmented (RAG) tools on medical-knowledge (MedQA), clinician-alignment (HealthBench), and
real-clinical-query benchmarks, with the clinical tools no better than a search engine's AI overview.
They also note, consistent with prior work~\citep{wu2024clasheval}, that retrieval can degrade a strong
base model. A natural question for the knowledge-graph community follows: does replacing flat document
retrieval with \emph{structured} graph grounding change the verdict, and---more fundamentally---under what
conditions does grounding help a capable LLM at all?

We study this with a graph+vector database (samyama-graph) over the public biomedical knowledge graph
PrimeKG~\citep{chandak2023primekg}, evaluating two grounding methods against an ungrounded baseline across
a ladder of base models. Our findings are negative on public data and sharply positive on out-of-training
data, and together they yield a simple, falsifiable account of when grounding helps. We deliberately
foreground reproduction and honest negatives: tagging a high-profile clinical result requires that every
claim survive scrutiny.

\paragraph{Contributions.}
\begin{enumerate}[leftmargin=1.5em,itemsep=2pt]
  \item \textbf{Reproduction \& grader calibration (\S\ref{sec:repro}).} We show the \citet{vishwanath2026general}
  HealthBench headline ($\sim$88) corresponds to the \emph{Consensus} variant, not full HealthBench
  (\texttt{oss\_eval}), where frontier answers and the dataset's own \emph{ideal} completions both score
  $\sim$46--47 under a grader calibrated to physician labels ($82.5\%$ agreement). We reproduce GPT-5.2
  Consensus $=90.9$ and MedQA $\approx 90$ (vs.\ reported $94.2$; model-version drift), and report a
  grader-parser bug that deflates scores if uncorrected.
  \item \textbf{A knowledge-boundary law for grounding (\S\ref{sec:exp}).} On PrimeKG, neither naive
  triple-RAG nor an agentic NLQ$\to$Cypher loop improves MedQA at any model strength; on a synthetic
  counterfactual KG the same pipeline lifts chance to near-perfect; and a \emph{hybrid} benchmark with
  both strata shows zero lift on known facts and a $+68$ to $+78$ point lift on novel facts \emph{within
  one experiment}. Grounding value is gated by whether the decisive fact is out-of-training.
  \item \textbf{A three-regime framing (\S\ref{sec:regimes})} aligning the experiments with the
  grounding-substrate view of \citet{kunkunuru2026assetops}: (i)~no-knowledge / direct LLM, (ii)~graph-aided
  (LLM queries the graph, optionally enriching it back as a chain of tasks), and (iii)~hybrid; with a
  \emph{no-LLM} deterministic arm that answers $100\%$ on both strata, isolating the data layer from the LLM.
  \item \textbf{A reusable agentic GraphRAG artifact (\S\ref{sec:method})} over samyama-graph (entity
  linking via vector search, self-correcting Cypher generation), plus two honest engine findings, and a
  pre-registration discipline that caught two spurious small-sample signals.
\end{enumerate}

\section{Background}\label{sec:bg}
\paragraph{Benchmarks.} MedQA~\citep{jin2021medqa} is USMLE-style multiple choice. HealthBench
\citep{arora2025healthbench} grades free-text answers against physician-written rubric items; it ships
multiple subsets, including the full \texttt{oss\_eval} set (many criteria per item, with negative-point
penalties) and a \emph{Consensus} subset (few, high-agreement, all-positive criteria). The scoring is a
points fraction $\text{score}=\operatorname{clip}\!\big(\sum_{\text{met}} p_i / \sum_{p_i>0} p_i,\,0,1\big)$,
which differs markedly between subsets.
\paragraph{Grounding.} GraphRAG retrieves subgraphs to ground generation; medical KG-RAG work typically
reports MedQA, while HealthBench augmentation has used text RAG~\citep{drinfo2026}. We use PrimeKG
\citep{chandak2023primekg} (drug/disease/phenotype/gene), and the samyama-graph engine for vector search
and Cypher. The agentic variant follows the knowledge-graph-as-data-layer thesis~\citep{kunkunuru2026assetops}:
the LLM \emph{queries} a deterministic graph rather than reading retrieved text.

\section{Three Regimes of Knowledge}\label{sec:regimes}
\begin{figure}[t]\centering
\resizebox{\linewidth}{!}{%
\begin{tikzpicture}[font=\footnotesize,>={Latex[length=2mm]},
  pth/.style={draw,rounded corners,align=left,text width=6.2cm,inner sep=4pt},
  eng/.style={draw,rounded corners,fill=blue!8,align=center,text width=2.9cm,inner sep=5pt,minimum height=26mm},
  ans/.style={draw,rounded corners,fill=green!10,align=center,text width=1.5cm,inner sep=4pt},
  ar/.style={->}]
\node[draw,rounded corners,fill=black!10,inner sep=4pt,align=center] (q) {Clinical\\question};
\node[pth,fill=red!6,right=12mm of q] (b) {\textbf{A\_agent (B).} vector entity-link $\to$ LLM writes Cypher $\to$ execute. \textit{lift $\approx$0 in-training; $+77$ out-of-training (82\% queries succeed)}};
\node[pth,fill=red!4,above=2mm of b] (at) {\textbf{A\_text} (flat RAG). source text in context. \textit{correct, but one LLM call \emph{per query}; unstructured}};
\node[pth,fill=gray!8,above=2mm of at] (a0) {\textbf{A0 — no knowledge.} LLM alone. \textit{works iff fact $\in$ training $\cup$ inferable (e.g.\ real drugs via INN nomenclature); else $\approx$chance}};
\node[pth,fill=orange!10,below=2mm of b] (bk) {\textbf{A\_kg (naive).} vector-search $+$ one-hop expand, \textbf{no agent}; format as triples. \textit{naive graph read; lift $\approx$0 in-training}};
\node[pth,fill=blue!6,below=2mm of bk] (c) {\textbf{A\_det (C).} deterministic Cypher, \textbf{no LLM}. \textit{100\% whenever the fact is graph-resident}};
\node[pth,fill=blue!6,below=2mm of c] (d) {\textbf{A\_GAK (D).} miss $\to$ agent extracts from a source $\to$ materialize (\texttt{provenance}) $\to$ cache. \textit{once per fact; novel $0\to100$}};
\node[eng,right=40mm of bk] (kg) {\mbox{\textbf{samyama-graph}}\\[2pt]\scriptsize HNSW vector $\cdot$ OpenCypher $\cdot$ graph algorithms $\cdot$ enrich/cache};
\node[ans,right=14mm of kg] (out) {Answer};
\foreach \n in {a0,at,b,bk,c,d} {\draw[ar] (q.east) -- (\n.west);}
\draw[ar] (a0.east) -| (out.north);
\draw[ar] (at.east) -| ([xshift=-3pt]out.north);
\draw[ar] (b.east) -- (kg.west);
\draw[ar] (bk.east) -- (kg.west);
\draw[ar] (c.east) -- (kg.west);
\draw[<->,dashed] (d.east) -- node[below,font=\scriptsize\itshape,align=center]{enrich (write)\\$+$ query (read)} (kg.west);
\draw[ar] (kg.east) -- (out.west);
\end{tikzpicture}}
\caption{All grounding paths over samyama-graph in one view (arms = paper3 architectures A/B/C/D). \emph{No
knowledge} (A0) succeeds only when the fact is in training or inferable from surface structure (e.g.\ drug
nomenclature); \emph{flat RAG} (A\_text) reads a source each query, while the naive KG-triple baseline
\textbf{A\_kg} (vector-search $+$ one-hop expand, no agent) is its graph-read analogue. The graph-aided paths --- LLM-written
Cypher (B), no-LLM deterministic (C), and generation-augmented knowledge (D: extract a novel fact from a
source, tag provenance, cache, then serve via C/B) --- make out-of-training facts answerable and are
redundant on in-training facts. Net: grounding helps iff the fact lies outside \emph{training $\cup$
inferable}.}
\label{fig:regimes}
\end{figure}
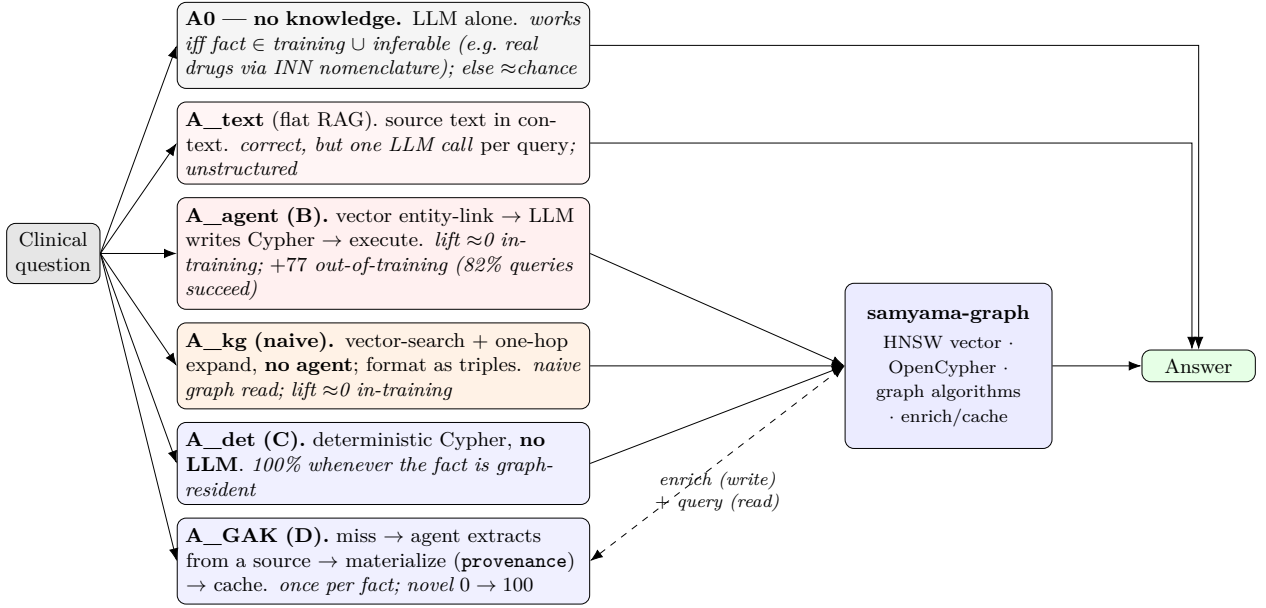

We organize the study around three regimes of where the decisive knowledge lives (Figure~\ref{fig:regimes}),
adapting the grounding-substrate view of \citet{kunkunuru2026assetops}:
\begin{enumerate}[leftmargin=1.5em,itemsep=1pt]
  \item \textbf{No knowledge (direct LLM).} The model answers from parametric memory alone (arm A0).
  \item \textbf{Graph-aided.} The graph is queried \emph{before} the LLM answers (NLQ$\to$Cypher; arm
  A\_agent) or \emph{computed without the LLM} at all (deterministic handler; arm A\_det), and---in the
  full substrate---missing facts can be written back and cached as a chain of tasks toward a goal
  (generation-augmented knowledge~\citep{kunkunuru2026assetops}).
  \item \textbf{Hybrid.} Some decisive facts are in the model's training \emph{and} in the graph (overlap,
  redundant) while others are only in the graph (novel, additive). This is the realistic deployment case.
\end{enumerate}
Our experiments instantiate all three: PrimeKG (\S\ref{sec:exp}) is the in-training overlap; the synthetic
KG is the pure out-of-training case; and the hybrid benchmark mixes both in one experiment.

\section{Reproduction and Grader Calibration}\label{sec:repro}
We reproduce the public components with deterministic decoding (temperature $0$, seed $62$) and the exact
selection procedure from the released harness (\texttt{random.Random(62).sample} over the source order,
single-user-turn filter), verified by matching the published filtered counts.

\paragraph{MedQA.} GPT-5.2 scores $90.0\%$ on the matched 500-item subset (100\% answer-parse rate), vs.\
the reported $94.2\%$. The gap is stable across two independent subsets and is unaffected by enabling web
search; we attribute it to model-version drift (the original was accessed four months earlier). This is a
faithful partial reproduction, not a discrepancy in method.

\paragraph{HealthBench: which scale?} On full \texttt{oss\_eval}, GPT-5.2 scores $46.5$ (macro) / $41.8$
(the released pooled formula). Crucially, the dataset's own \emph{ideal} completions---the best possible
answers---score only $\sim$47 under our grader. No model can score $88$ when the ideal answer scores $47$;
the headline must use a different scale. The \emph{Consensus} variant (few, all-positive criteria) does
reach it: GPT-5.2 Consensus $=90.9$ (95\% CI $[87.4,94.4]$, $N{=}150$), bracketing the reported $88.0$
(Table~\ref{tab:repro}).

\paragraph{Grader calibration.} Our judge panel (gpt-4.1 + Claude Opus 4.6) agrees with physician
gold labels on HealthBench's meta-evaluation set at $82.5\%$, matching the original HealthBench grader's
reported agreement; thus the $\sim$47 ideal-completion score reflects the benchmark's intended difficulty,
not a harsh grader. We also document a parser failure mode---conflating ``criteria not met'' with a
parse failure---that silently deflated scores until hardened (it returns true/false/none distinctly).

\begin{table}[t]\centering\small
\caption{Reproduction. The Nature headline HealthBench number is the Consensus scale, not full
\texttt{oss\_eval}. Ideal completions bound the full-set scale at $\sim$47.}
\label{tab:repro}
\begin{tabular}{lccc}
\toprule
Setting & GPT-5.2 & Ideal completions & Reported \\
\midrule
HealthBench \texttt{oss\_eval} (full) & 46.5 / 41.8 & $\sim$47 & --- \\
HealthBench \emph{Consensus} & \textbf{90.9} \,[87.4, 94.4] & $\sim$82 & \textbf{88.0} \\
MedQA (matched 500) & 90.0 & --- & 94.2 \\
\bottomrule
\end{tabular}
\end{table}

\section{Method: grounding over samyama-graph}\label{sec:method}
We load PrimeKG into samyama-graph (an OpenCypher graph+vector engine) and embed node names for HNSW
vector search. We compare the arms below, all with a fixed base model: A0 is the ungrounded baseline; A\_kg and A\_text are
naive grounding (naive KG triples vs.\ raw source text); A\_agent, A\_det, and A\_GAK are the graph paths of
Figure~\ref{fig:regimes}.
\begin{itemize}[leftmargin=1.5em,itemsep=1pt]
  \item \textbf{A0}: base model, no retrieval.
  \item \textbf{A\_kg (naive KG-RAG)}: vector-search the question, expand a one-hop clinical subgraph (degree-capped),
  format as triples.
  \item \textbf{A\_text (flat text-RAG)}: the relevant source passage is placed directly in the LLM context,
  with no graph; one LLM call per query.
  \item \textbf{A\_agent (agentic)}: a fixed agent (gpt-4.1) does entity linking via vector search,
  writes multi-hop Cypher constrained to verbatim resolved entity names and specific relationship types,
  executes it on samyama-graph, and self-corrects for up to two rounds on error or empty result; the
  returned rows are the grounding context.
  \item \textbf{A\_det (deterministic, no LLM)}: a pre-coded handler issues the Cypher lookup directly
  (no LLM in the loop) and matches the result to an option. This isolates the \emph{data layer} from the
  LLM and mirrors Architecture~C of \citet{kunkunuru2026assetops}.
  \item \textbf{A\_GAK (generation-augmented knowledge)}: the realistic out-of-training loop
  (Architecture~D of \citet{kunkunuru2026assetops}). The decisive fact lives in an unstructured
  \emph{source} (not in training, not yet in the graph); on a miss, the agent extracts it and materializes
  provenance-tagged nodes/edges into the graph, which is then queried deterministically. Repeat questions
  about the same entity are \emph{cache hits} (no LLM).
\end{itemize}

\paragraph{Scope of engine capabilities used.} A\_agent/A\_det exercise the engine's vector search and
Cypher; we do \emph{not} use its graph-algorithm (PageRank, paths) or optimization (NSGA-II) primitives,
because clinical multiple-choice is lookup-shaped. A different question class (criticality, interaction
paths, treatment scheduling) would route to those Tier-2 primitives; we leave that to future work.
Fixing the query-writer isolates whether the \emph{answering} model benefits from grounding rather than
its Cypher skill. The agent reaches an $82\%$ non-empty query rate on MedQA after vector entity-linking
and few-shot direction examples (from $0\%$ when the LLM invented entity names).

\paragraph{Engine findings (honest), and what became of them.} Three engine gaps shaped
the implementation reported here. (i) Cypher \texttt{expand} lacked LIMIT-pushdown on
high-degree nodes, so an unbounded one-hop expansion over hub entities materialised
millions of rows; we work around it with a precomputed degree-capped adjacency.
(ii) \texttt{type(r)} over multi-type or optional relationships errored (``type()
requires an edge''). (iii) A comma-separated \texttt{CREATE} that declares nodes then
references their variables for edges materialised nodes but not edges, so GAK issues
node-creates then edge-creates via \texttt{MATCH}.

\textbf{All three are fixed in the engine's v1.8.0 release} (2026-09-15), measured
2026-09-21 on that release: \texttt{type(r)} returns the edge types for a multi-type
pattern and \texttt{null} for an unmatched optional relationship rather than erroring;
the comma-separated \texttt{CREATE} materialises the edge; and \texttt{LIMIT}~5 over a
degree-20{,}000 hub returns in 1.1\,ms against 1.3\,ms over a degree-1{,}000 hub, so the
expansion no longer scales with degree. The experiments in this paper were run on the
engine as it then was and are unchanged --- what changes is that the workarounds they
required are no longer necessary.

\section{Experiments}\label{sec:exp}
\paragraph{Setup.} MedQA exact-match, $N{=}150$--$200$, base-model ladder gpt-4.1-nano $\to$ gpt-4o-mini
$\to$ gpt-4.1 $\to$ gpt-5.2 (weak$\to$strong). Per-question retrieval is computed once and reused across
base models; intervals are 95\% Wilson/normal.

\paragraph{Public KG: no lift.} Table~\ref{tab:public} shows neither naive nor agentic grounding moves
MedQA accuracy at any model strength; every lift is within $\pm3.4$ points and inside the confidence
intervals, despite the agent issuing correct queries $82\%$ of the time (it does return correct facts,
e.g.\ retrieving first-line antibiotics for a diphtheria vignette). A pilot at $N{=}8$ showed a clean
monotone ``crossover'' ($+25\to-12.5$); it vanished at $N{=}200$---a reminder that small-$N$ grounding
signals are noise, which our pre-registration was designed to catch.

\begin{table}[t]\centering\small
\caption{Public KG (PrimeKG): MedQA accuracy. Naive and agentic grounding give no significant lift at any
model strength (CIs $\pm4$--$7$).}
\label{tab:public}
\begin{tabular}{lccccc}
\toprule
Base model & A0 & A\_kg & A\_agent & $\Delta_{\text{naive}}$ & $\Delta_{\text{agent}}$ \\
\midrule
gpt-4.1-nano & 71.3 & 69.3 & 72.7 & $-2.0$ & $+1.4$ \\
gpt-4o-mini  & 82.7 & 79.3 & 82.0 & $-3.4$ & $-0.7$ \\
gpt-4.1      & 89.3 & 90.0 & 90.0 & $+0.7$ & $+0.7$ \\
gpt-5.2      & 92.7 & 91.3 & 92.0 & $-1.4$ & $-0.7$ \\
\bottomrule
\end{tabular}
\end{table}

\paragraph{Out-of-training KG: decisive.} We build a synthetic counterfactual clinical KG (fictional
drugs, diseases, phenotypes; seed $62$) whose facts cannot be in any model's training, and pose
KG-answerable multiple-choice questions. Table~\ref{tab:synth}: A0 sits at chance ($\sim$22\%, four
options), while the \emph{identical} agentic pipeline reaches $\sim$100\% ($+75$ to $+79$, $100\%$ query
success), uniformly across the ladder---because all models equally lack the knowledge.

\begin{table}[t]\centering\small
\caption{Out-of-training KG (synthetic counterfactual): the same agentic samyama-graph grounding turns
chance into near-perfect at every model strength ($N{=}150$).}
\label{tab:synth}
\begin{tabular}{lccc}
\toprule
Base model & A0 (alone) & A\_agent (graph) & $\Delta_{\text{agent}}$ \\
\midrule
gpt-4.1-nano & 23.3 & 98.7 & $+75.4$ \\
gpt-4o-mini  & 22.0 & 100.0 & $+78.0$ \\
gpt-4.1      & 23.3 & 100.0 & $+76.7$ \\
gpt-5.2      & 21.3 & 100.0 & $+78.7$ \\
\bottomrule
\end{tabular}
\end{table}

\begin{figure}[t]\centering
\includegraphics[width=0.92\linewidth]{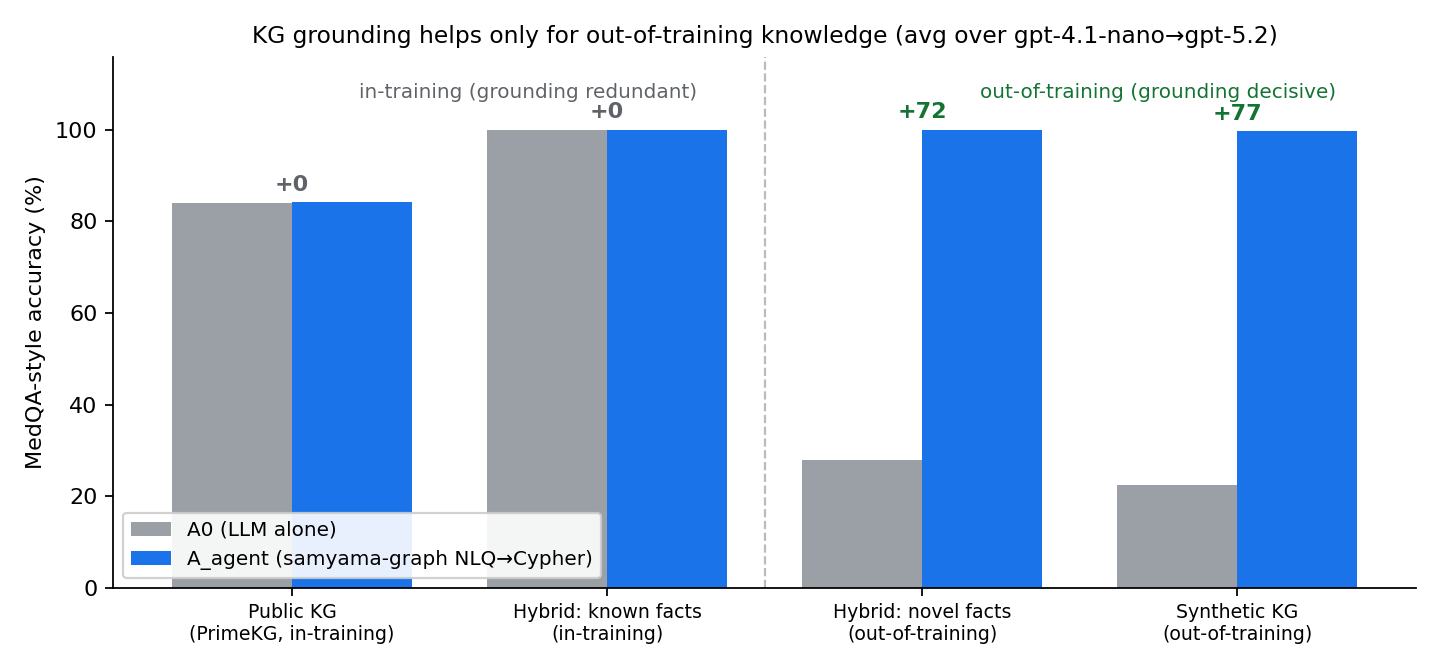}
\caption{Across all settings: KG grounding (A\_agent) matches A0 on in-training knowledge (lift $\approx0$)
and is decisive on out-of-training knowledge (lift $+72$ to $+77$), averaged over the gpt-4.1-nano$\to$gpt-5.2
ladder.}
\label{fig:results}
\end{figure}

\paragraph{Hybrid: the boundary within one benchmark.} The cleanest test holds everything else fixed and
varies only fact-novelty \emph{across items of one benchmark}. We build a hybrid KG with a \emph{known}
stratum (24 canonical textbook disease$\to$drug facts the models reliably know) and a \emph{novel} stratum
(80 synthetic counterfactual facts), and load both into one graph. Table~\ref{tab:hybrid}: on the known
stratum A0 is already $100\%$ and grounding adds nothing ($+0$); on the novel stratum A0 is at chance
($\sim$22--32\%) and agentic grounding lifts it to $100\%$ ($+68$ to $+78$). The no-LLM A\_det handler
answers $100\%$ on \emph{both} strata---whenever the fact is graph-resident, the data layer alone suffices.
Figure~\ref{fig:results} summarizes all settings: grounding is flat in-training and decisive out-of-training.

\begin{table}[t]\centering\small
\caption{Hybrid benchmark (one experiment, two strata). Lift is concentrated entirely on novel
(out-of-training) facts; A\_det (no LLM) is 100\% on both.}
\label{tab:hybrid}
\begin{tabular}{lcccc}
\toprule
 & \multicolumn{2}{c}{Known stratum (in-training)} & \multicolumn{2}{c}{Novel stratum (out-of-training)} \\
\cmidrule(lr){2-3}\cmidrule(lr){4-5}
Base model & A0 & A\_agent & A0 & A\_agent \\
\midrule
gpt-4.1-nano & 100 & 100 & 22.5 & 100 \\
gpt-4o-mini  & 100 & 100 & 25.0 & 100 \\
gpt-4.1      & 100 & 100 & 32.5 & 100 \\
gpt-5.2      & 100 & 100 & 31.2 & 100 \\
\midrule
A\_det (no LLM) & \multicolumn{2}{c}{100} & \multicolumn{2}{c}{100} \\
\bottomrule
\end{tabular}
\end{table}

\paragraph{GAK: the mechanism that puts novel facts in the graph.} The synthetic and hybrid experiments
load novel facts into the graph by construction; the realistic loop must \emph{get them there}. We test
generation-augmented knowledge (A\_GAK) on 40 novel facts (80 questions) whose answer lives only in an
unstructured source monograph. Starting from an empty graph: A0 (no source) is at chance ($20\%$); flat
text-RAG (source in context) and A\_GAK both reach $100\%$ (Table~\ref{tab:gak}). The difference is
economics and form: text-RAG pays an LLM call \emph{per query} ($80$), whereas A\_GAK extracts each fact
\emph{once} ($40$ enrichment calls), materializes it provenance-tagged (all $160$ nodes
\texttt{source:enriched}), and serves the remaining $40$ questions as deterministic cache hits (no LLM)---
asymptotically zero marginal cost, queryable, and auditable. This closes the loop: novel knowledge enters
the graph via the agent, then behaves like the in-graph facts of \S\ref{sec:exp}.

\begin{table}[t]\centering\small
\caption{A\_GAK (Architecture D): facts live only in unstructured sources, absent from training and the
graph. Accuracy parity with text-RAG, but the LLM is paid once per fact (then cached), with provenance.}
\label{tab:gak}
\begin{tabular}{lcccc}
\toprule
Arm & Accuracy & LLM calls (80 q) & Cache hits & Provenance-tagged nodes \\
\midrule
A0 (no source) & 20.0 & 80 & --- & --- \\
A\_text (flat RAG) & 100.0 & 80 & --- & --- \\
A\_GAK (enrich+cache) & 100.0 & \textbf{40} & \textbf{40} & 160 (\texttt{source:enriched}) \\
\bottomrule
\end{tabular}
\end{table}

\paragraph{Real out-of-training facts, and a refinement: ``out-of-training'' $\ne$ ``unknown''.}
We test 13 \emph{real} 2026 FDA novel-drug approvals (verified; post-dating most ladder models), asking
``which drug is indicated for condition $X$?'' with other recent drugs as distractors. Surprisingly,
A0~$\approx84\%$ (per-drug hit-rate $50$--$100\%$ across the ladder, with run-to-run noise)---the models \emph{infer} most indications from International Nonproprietary Name (INN)
nomenclature (\texttt{-drostat}$\to$hypertension; \texttt{-penem}$\to$UTI; \texttt{-trelvir}$\to$COVID;
\texttt{copper}$\to$Menkes).
A post-cutoff drug is therefore not automatically out of reach: its \emph{name carries class information}.
The lone genuine isolation is the one approval whose name encodes nothing about its indication---blastic
plasmacytoid dendritic cell neoplasm (pivekimab sunirine): A0 \emph{reliably fails} it ($\approx$$8\%$,
$1/12$ over repeated trials) while A\_det (deterministic graph lookup) and A\_agent answer it correctly.
Even nominally opaque names are usually inferred---A0 still gets Hunter syndrome (tividenofusp alfa)
$\approx$$83\%$ and achondroplasia (navepegritide) $\approx$$50\%$ of the time---which sharpens rather
than softens the point. This refines the law:
\textbf{grounding helps to the extent the fact is neither in training \emph{nor} reconstructable from
surface structure} (boundary $=$ training $\cup$ inferable). The synthetic KG (random names, no stems) is
the pure isolation; real nomenclature is itself latent knowledge, and grounding (deterministically, via
A\_det) still guarantees the uninferable residual.

\paragraph{Robustness.} Beyond the within-run 95\% CIs, the out-of-training effect is stable across
KG-generation seeds: the synthetic lift is $76.6\pm2.9$ over 3 seeds $\times$ 4 models (range $[72,82]$),
and the hybrid novel-stratum lift is $\sim$$77\pm8$. The public-KG null is stable across two independent
full runs (all $|\Delta|\le5.5$ both times). Generation is deterministic (temperature $0$, seed $62$);
the headline conclusions do not hinge on a single run.

\FloatBarrier
\section{Discussion: a knowledge-boundary law}\label{sec:disc}
The two regimes differ only in the \emph{novelty} of the grounding facts---same engine, same agentic
pipeline, same models---and that alone flips the result from null to decisive. We state the account
plainly: structured grounding raises accuracy only to the extent it supplies information the model's
parametric prior lacks. Public-KG facts (PrimeKG is in the training corpora) are redundant; even when the
graph returns a correct fact, the model already knew it. Out-of-training facts are not, so grounding is
decisive. This is consistent with the prior-versus-evidence view of retrieval~\citep{wu2024clasheval} and
explains why public-benchmark evaluations \emph{understate} the value of grounding for the deployment that
matters: private and institutional data. It operationalizes \citet{vishwanath2026general}'s own
observation that the path forward lies with hospital-specific data, and the knowledge-graph-as-data-layer
thesis of \citet{kunkunuru2026assetops}.

\section{Related Work}
Medical KG-RAG systems report MedQA gains but rarely isolate the knowledge boundary; HealthBench
augmentation has used text RAG~\citep{drinfo2026}; MIRAGE/MedRAG~\citep{xiong2024mirage} benchmarks
medical RAG. Agentic and budgeted graph retrieval~\citep{he2024gretriever,agrag2025} provide the
machinery we adopt rather than claim. ClashEval~\citep{wu2024clasheval} frames the prior-versus-evidence
tension we explain via novelty. Health-system-scale models~\citep{jiang2023healthsystem} and the source
study~\citep{vishwanath2026general} motivate the institutional-data direction our synthetic result makes
precise.

\section{Limitations and Honest Negatives}\label{sec:lim}
The headline of our grounding study is a \emph{negative} on public data; we report it because the agentic
pipeline was built well ($82\%$ query success, verifiably correct facts) and still did not help. The
synthetic positive uses fictional entities; it cleanly isolates the mechanism but is not a clinical
benchmark---demonstrating the predicted lift on a \emph{real} private clinical graph is future work. The
crossover predicted by a prior-strength account is not visible here because both regimes are
endpoints (prior fully covers, or is absent); an intermediate, graded-prior task would test it. Query
generation uses a fixed capable agent (gpt-4.1); a weak model writing its own Cypher would conflate
querying and answering skill. Grader and reproduction caveats are in \S\ref{sec:repro}. MedQA exact-match
ignores reasoning quality; clinical tools in the source study lack APIs and are cited, not re-run.

\section{Conclusion}
Structured KG grounding does not beat frontier LLMs on public medical benchmarks---not because grounding
is weak, but because public-KG facts are already in the models. The same agentic samyama-graph pipeline is
decisive when the knowledge is out-of-training. The actionable claim for practitioners is to ground on
data the model has not seen---private, recent, institutional---and to evaluate there, since public
benchmarks understate grounding's value. Along the way we corrected the scale of a widely-cited HealthBench
result and surfaced two fixable engine gaps. Code and one-command reproduction are open.

\paragraph{Reproducibility.} All numbers regenerate from the released harness
(\url{https://github.com/samyama-ai/clinical-llm-graphrag}): deterministic decoding (seed $62$), committed
subset hashes, the synthetic KG generator (seed $62$), and the agentic pipeline. Public data: MedQA,
HealthBench (CC-BY-4.0), PrimeKG (MIT).

\small
\bibliographystyle{plainnat}
\bibliography{paper15_clinical_llm_graphrag}
\end{document}